\documentclass[sigconf]{acmart}

\graphicspath{{figs/}}

\usepackage{booktabs} 

\copyrightyear{2018} 
\acmYear{2018} 
\setcopyright{acmcopyright}
\acmConference[MM '18]{2018 ACM Multimedia Conference}{October 22--26, 2018}{Seoul, Republic of Korea}
\acmBooktitle{2018 ACM Multimedia Conference (MM '18), October 22--26, 2018, Seoul, Republic of Korea}
\acmPrice{15.00}
\acmDOI{10.1145/3240508.3240674}
\acmISBN{978-1-4503-5665-7/18/10}

\begin{document}
\title{Support Neighbor Loss for Person Re-Identification}


\author{Kai Li$^{1}$, Zhengming Ding$^{2}$, Kunpeng Li$^{1}$, Yulun Zhang$^{1}$ and Yun Fu$^{1,3}$}
\affiliation{%
  \institution{$^1$ Department of Electrical and Computer Engineering, Northeastern University, Boston, USA}
  \institution{$^2$ Department of Computer, Information and Technology, Indiana University-Purdue University} 
  \institution{$^3$ College of Computer \& Information Science, Northeastern University, Boston, USA} 
}
\email{{kaili,kunpengli,yunfu}@ece.neu.edu, zd2@iu.edu, yulun100@gmail.com}

\renewcommand{\shortauthors}{K. Li et al.}

\begin{abstract}
Person re-identification (re-ID) has recently been tremendously boosted due to the advancement of deep convolutional neural networks (CNN). The majority of deep re-ID methods focus on designing new CNN architectures, while less attention is paid on investigating the loss functions. Verification loss and identification loss are two types of losses widely used to train various deep re-ID models, both of which however have limitations.  Verification loss guides the networks to generate feature embeddings of which the intra-class variance is decreased while the inter-class ones is enlarged. However, training networks with verification loss tends to be of slow convergence and unstable performance when the number of training samples is large. On the other hand, identification loss has good separating and scalable property. But its neglect to explicitly reduce the intra-class variance limits its performance on re-ID, because the same person may have significant appearance disparity across different camera views. To avoid the limitations of the two types of losses, we propose a new loss, called support neighbor (SN) loss. Rather than being derived from data sample pairs or triplets, SN loss is calculated based on the positive and negative support neighbor sets of each anchor sample, which contain more valuable contextual information and neighborhood structure that are beneficial for more stable performance. To ensure scalability and separability, a softmax-like function is formulated to push apart the positive and negative support sets. To reduce intra-class variance, the distance between the anchor's nearest positive neighbor and furthest positive sample is penalized. Integrating SN loss on top of Resnet50, superior re-ID results to the state-of-the-art ones are obtained on several widely used datasets.






\end{abstract}

%
%

\copyrightyear{2018} 
\acmYear{2018} 
\setcopyright{acmcopyright}
\acmConference[MM '18]{2018 ACM Multimedia Conference}{October 22--26, 2018}{Seoul, Republic of Korea}
\acmBooktitle{2018 ACM Multimedia Conference (MM '18), October 22--26, 2018, Seoul, Republic of Korea}
\acmPrice{15.00}
\acmDOI{10.1145/3240508.3240674}
\acmISBN{978-1-4503-5665-7/18/10}



\maketitle

\textbf{ACM Reference Format:}
Kai Li, Zhengming Ding, Kunpeng Li, Yulun Zhang, and Yun Fu. 2018. Support Neighbor Loss for Person Re-Identification. In 2018 ACM Multimedia Conference (MM' 18), October 22-26, 2018, Seoul, Republic of Korea. ACM, New York, NY, USA, 9 pages. https://doi.org/10.1145/3240508.3240674

\section{Introduction}
Person re-identification (re-ID) has been attached with increasing attentions in recent years due to its importance for many real-world applications, such as video surveillance, robotics, human-computer interaction, etc. 
Given an image of the person of interest, the goal of re-ID is to find the other images of the same person captured by different cameras or the same cameras in different time.
By the nature of this task, two fundamental problems need to be solved.  The first is to generate representations that encode the most discriminative appearance cues of a person, such that his/her representations from different camera views are similar.
The second is to develop appropriate distance metrics under which images of the same person are more similar than those of different persons in terms of the representation distance.

Conventional methods consider the two problems separately. Some focus on developing robust pedestrian image descriptor \cite{DBLP:conf/eccv/MaSJ12,ZhaoOW13,wu2015viewpoint,Li2018Discriminative}, 
while others strive to develop effective distance metrics \cite{DBLP:conf/cvpr/MignonJ12,DBLP:conf/cvpr/PedagadiOVB13,XiongGCS14,liao2015person}.
Benefiting from the power of the deep convolutional neural network (CNN) in generating discriminative representations, as well as the availability of large-scale annotated datasets, the re-ID community has witnessed a rapid blossom in the most recent several years \cite{li2017learning,bai2017scalable,zhao2017deeply,zheng2017unlabeled,su2017pose}. Unlike the conventional algorithms, deep learning based re-ID methods learn from data feature representations and distance metrics jointly in an end-to-end manner. 


\begin{figure*}
\begin{center}
\includegraphics[width=0.9\linewidth]{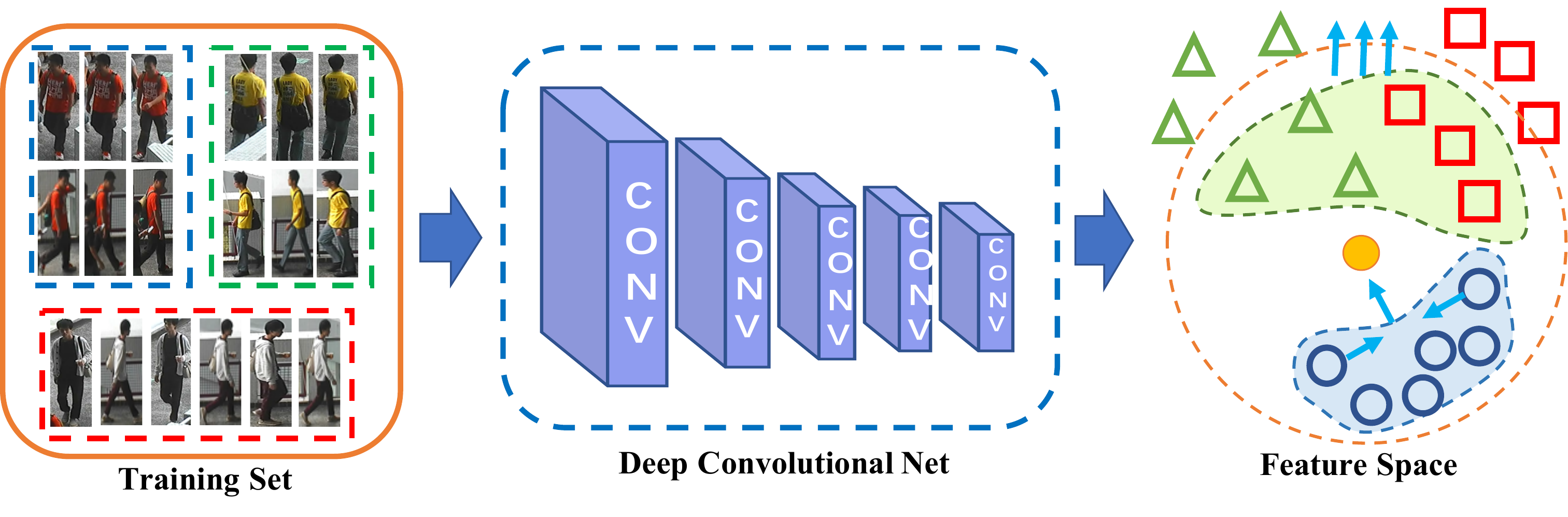}
\end{center}
\vspace{-0.15in}
\caption{Framework of the proposed method. A batch of images are fed to a deep neural network to get the feature embeddings. 
Taking each sample as the anchor in the embedding space, we calculate its $K$-NN neighbors and
pull the positive neighbors towards it while push the negative ones away from it.
Meanwhile, the positive neighbors are squeezed together to form a compact cluster. 
} 
 \vspace{-0.1in}
	\label{framework}		
\end{figure*}

The majority of deep re-ID methods focus on designing new CNN architectures to model the pose variance, human body misalignment, occlusion, etc \cite{chen2017multi,li2017person,zheng2017discriminatively,ahmed5improved,varior2016gated}. However, much less effort has been paid on the investigation of the loss.
Two types of losses are extensively utilized for re-ID, i.e., verification loss and identification loss.
Verification loss strives to reduce the intra-class variance while enlarge the inter-class one.
This property makes verification loss a seemingly natural choice for re-ID, since the same person can have substantial appearance disparity owing to the viewpoint, pose and background variations. 
Deep neural networks which model the large intra-class variations shall be promising to output good verification results during the test time.
Despite of the success in a number of re-ID algorithms \cite{cheng2016person,varior2016gated,xiao2016learning,zheng2017discriminatively}, verification loss based methods shall suffer the common problem that 
they are prone to show slow convergence and unstable performance in the circumstance of numerous person identities. 
Some well designed hard mining strategy can somewhat mitigate this problem \cite{hermans2017defense,song2016deep}.
But it is ineradicable since analyzing individual pairs or triplets of a example does not employ sufficient contextual insight of the neighborhood structure, thereby leading the model fails to learn the normal associations of the person. 

On the other hand, identification loss has good separation and scalable property, and it is extensively used in large-scale classification problem \cite{he2016deep}. When applying identification loss for re-ID, unlike verification loss based methods where the similarity metric is directly encoded as the training objective, a classifier is trained on the training identities and nearest neighbor query is performed at test time using the feature representations of the trained network \cite{xiao2016learning,zheng2017person}. Identification loss has good property of separating different classes, but it does not explicitly encode the intra-class variance, thus shall cause unsatisfactory results for re-ID where the intra-person variance could be significant.
Some methods \cite{chen2017multi,li2017person,zheng2017discriminatively} attempt to combine these two types of losses in multi-task fashion. 
Typically, the networks of these methods often consist of multiple branches, with each branch corresponding to one type of loss.
How to balance different branches remains a challenging problem, and the multi-branch structures definitively make the models hard to train and less efficient for inference.

Instead, we propose a new loss, called support neighbor (SN) loss, which avoids the problems of both identification and verification losses. 
Taking each sample within a batch as the anchor, we calculate its $K$-NN neighbors and define the positive samples and negative samples among the $K$-NN neighbors as \textbf{support neighbors}. SN loss is calculated based on the support neighbors. 
In order to encourage intra-class cohesion, 
we penalize the difference of the distances of the furthest positive neighbor and the nearest positive neighbor to the anchor. 
To ensure class separation, we formulate a softmax-like function to maximize the similarity of positive neighbors to the anchor and meanwhile minimize the similarity of negative neighbors to the anchor.

The support neighbors of a sample are the most similar samples of the anchor within the batch, thereby containing the most important information of the person the anchor image captures. Unlike triplet, which contains very limited contextual information and neighborhood structure, our SN loss is exposed with more such valuable information, so deep models learned based on our SN loss can encode more discriminative information about persons. We verify the effectiveness of the proposed SN loss by integrating it on top of the common feature extraction network Resnet50 \cite{he2016deep}. The experiments show we achieve remarkable improvements over the state-of-the-art methods on several commonly benchmarking datasets. Figure \ref{framework} shows the framework of the proposed method.

\section{Related Works}
\label{related_works}
Person re-ID is a very hot topic in recent years and numerous algorithms have been proposed.
Here we give a brief overview of methods in this area, with an inclination of deep learning based methods.
Zheng \textit{et al} had conducted a comprehensive survey; we refer interested readers to their paper \cite{zheng2016person}. 

Conventional re-ID methods focus either on pedestrian image description or distance metric learning.
Feature generation based methods underlie on the fact that images of the same person in different camera views should be similar in appearance.  Since a single image descriptor (\textit{i.e.}, RGB, LBP, SIFT, etc.) is often not powerful enough to encode all the information that are essential for pedestrian image matching, concatenating the feature vectors of several image descriptor is commonly used~\cite{ZhaoOW13-2,li2013learning,chen2016similarity,shen2015person,matsukawa2016hierarchical}. Besides directly using low-level color and texture features, 
some methods also strive to utilize pedestrian attributes which are     
more robust to image transformations \cite{layne2012person, liu2012attribute, su2015multi}. 

Due to the high-dimensional nature of pedestrian image features, it is critical to learn a good distance metric to obtain the invariant factors among sample variances. The general idea of metric learning based re-ID methods is to learn some distance metrics under which the vectors of the same identities are pushed closer while the vectors of different identities are pulled further apart. The most acknowledged metric learning based person re-ID algorithm is KISSME~\cite{DBLP:conf/cvpr/KostingerHWRB12}, which decides whether a pair of description vectors is similar or not by formulating it as a likelihood ratio test under the 
assumption that the feature distances obey a Gaussian distribution with a zero mean. Inspired by KISSME, many metric learning based person re-ID algorithms have been proposed, including LFDA~\cite{XiongGCS14}, XQDA~\cite{liao2015person}, MLAPG~\cite{liao2015efficient}.



Deep learning based methods have tremendously pushed forward the boundary of re-ID. These methods focus on designing various deep CNN structures to learn discriminative feature embeddings
and/or strive to devise better loss functions for training the networks.
To model the pose variations and viewpoint changes in cross-view pedestrian images, a number of methods introduce some special units into the architectures. 
Ahmed \textit{et al.} \cite{ahmed5improved} proposed to capture local relationship between two images via
a cross-input neighborhood difference layer and designed a patch summary layer to summarize the features obtained in the previous layers. Varior \textit{et al.} \cite{varior2016gated} proposed to use a gating function to compare features along a horizontal stripe and output a gating mask to indicate importance of the local patterns. 
To solve the misalignment of human bodies in different images, some methods design multi-branch model architectures, with each branch corresponding to a body part or the whole bodies, and fuse all branches together as the final feature embedding \cite{li2017learning,chen2017multi}.
Some methods introduce attention modules to emphasize the most discriminative parts of human body. Liu \textit{et.al} \cite{liu2017end} built a recurrent Siamese neural network which generated attention maps according to the comparison over triplets of person images. Zhao \textit{et al.} \cite{zhao2017deeply} designed a network which jointly modeled representation computation and body part extraction in an end-to-end fashion by maximizing the re-identification quality. 

Compared with designing various deep CNN architectures, less attentions have been attached on the loss function. The majority of the existing methods employ existing verification loss to train their models, including triplet loss \cite{cheng2016person,wang2016joint} contrastive loss \cite{varior2016siamese,varior2016gated}, binary identification loss \cite{li2014deepreid,ahmed5improved}. Recently, some seminal investigation has been made on improving the loss functions. Some approaches \cite{xiao2016learning,zheng2017person} attempt to use identification loss to learn discriminative features efficiently.
The combination of both verification and identification losses has also been initialized 
in \cite{chen2017multi,li2017person,zheng2017discriminatively}.  Hermans \textit{et al.} \cite{hermans2017defense} recently proposed a new formulation of calculating triplet loss and mitigated the limitation of conventional triplet loss calculation mechanism.
Chen \textit{et al.} \cite{chen2017beyond} proposed the quadruplet loss which added one more negative sample into the triplet and reached better generalization ability.

Our method also focuses on the loss function, but unlike existing methods which investigate pairs, triplets, or quadruplets of anchor samples, we investigate the neighbors of anchors, which require minimal data sampling and involve more contextual information, thereby securing stable and superior performance.

\section{Algorithm}
\label{algorithm}
This section will introduce the proposed Support Neighbor (SN) loss in details.
Before elaborating it, we will 
first give a brief introduction of softmax loss, the most popular identification loss; and triplet loss, the most widely used verification loss.
We will analyze their limitations and show how they motivate us to develop the proposed SN loss.

Given a training dataset $\mathcal{S}=\{(\mathbf{I}_i, y_i)\}_{i=1}^N$ of $N$ pedestrian images of $M$ identities, 
where $\mathbf{I}_i$ is the $i$-th pedestrian image, and $y_i$ is its corresponding identity label. The goal of (supervised) deep learning based re-ID methods is to learn from $\mathcal{S}$ a feature embedding model $\Omega_{\theta}$, parameterized by $\theta$. 
With $\Omega_{\theta}$, 
the feature embeddings of any query image $\textbf{I}_q$ and the gallery images 
$\mathcal{G}=\{\textbf{I}^j_g\}^G_{j=1}$ 
can be obtained as $\textbf{x}_q=\Omega_{\theta}(\textbf{I}_q)$ and 
$\mathcal{G}'=\{\textbf{x}^j_g\}^{G}_{j=1}$, where
$\textbf{x}^j_g = \Omega_{\theta}(\textbf{I}^j_g)$.
Then, the similarity of $\textbf{x}_q$ and every $\textbf{x}^j_g\in\mathcal{G}'$ is calculated, and the most similar one(s)
are return as the re-ID results.
To learn $\Omega_{\theta}$, some loss function 
$L(\theta)=f(\{\textbf{x}_i, y_i\}^N_{i=1})$ is to be minimized, where $\textbf{x}_i=\Omega_{\theta}(\textbf{I}_i)$.








\subsection{Existing Losses}




\noindent\textbf{Softmax loss}.
Softmax loss is widely used for multi-class classification task in which the goal is to enlarge the inter-class difference:
\begin{equation}
\begin{array}{cl}
L_{stm}(\theta)= -\frac{1}{N}\sum_{i=1}^N \log\frac{\exp(\mathbf{W}^\textrm{T}_{y_i}x_i+b_{y_i})}
{\sum_{j=1}^M \exp(\textbf{W}_j^{\textrm{T}}x_i+b_j)},
\end{array}
\label{softmax_loss}
\end{equation}
where $\textbf{W}_j$ and $\textbf{b}_j$ are $j$-th column of the weight and bias matrices.
Minimizing Eq.~\eqref{softmax_loss} with training dataset $\mathcal{S}$, $\Omega_{\theta}$ can be obtained.  
With $\Omega_{\theta}$, the embeddings of the query images and gallery images can be obtained. 
Then, the nearest neighbor(s) of $\textbf{x}_q$
among $\mathcal{G}'=\{\textbf{x}^j_g\}_{j=1}^G$ are regarded as the retrieval results.


One may have noticed that softmax loss only aims to ensure separation of different classes, but neglects to decrease the intra-class variance, which however is crucial for re-ID because the same person could be captured of great appearance disparity in different camera views. Thus, simply using softmax loss for re-ID may not lead to satisfactory results.

\noindent\textbf{Triplet loss}.
Verification loss seems more suitable for re-ID because it explicitly penalizes intra-class variance, while encourages the inter-class one.
Taking the famous triplet loss as an example, its objective function is 
\begin{equation}
\begin{array}{cl}
L_{tri}(\theta) = \sum_{i=1}^N \big[\|\textbf{x}_i-\textbf{x}_p\|_2^{2}-\|\textbf{x}_i-\textbf{x}_n\|_2^2+\alpha\big]_{+},
\end{array}
\label{triplet_loss}
\end{equation}
where $\alpha$ is the distance margin, $\textbf{x}_i$, $\textbf{x}_p$, and $\textbf{x}_n$ are the anchor, positive sample and negative
sample, respectively.

The foremost limitation of triplet loss is that the number of triplets is cubical to the number of samples,
which shall significantly slow down model convergence and lead to unstable performance.
Some hard mining strategies \cite{ding2015deep,shi2016embedding} can somewhat mitigate this problem.
However, regardless of which type of mining is being done, it is a separate step from training and adds considerable overhead, as it requires embedding a large fraction of the data with the most recent model and computing all pairwise distances between those data points.

Recently, Hermans \textit{et al.} \cite{hermans2017defense} proposed a novel batch sampling strategy, which avoids performing triplet construction on the whole dataset, but on mini-batches, thereby significantly improving the training efficiency and performance stableness.
The main idea is to randomly select a constant number of person with a constant number of images for each person in each batch. 
Within each batch, two variants of triplet losses are formulated. The first one is the Batch Hard (BH) variant, 
which selects the hardest triplets within the batch; the second one is the Batch All (BA) variant, which permutes all triplets within the batch.
Though being proved effective, both the two variants have limitations. For the BH variant, it selects only one triplet for an anchor sample, 
risking that the typical patterns of the identity fail to be encoded.
On the other hand, the BA variant permutes all possible positive-negative sample combinations, and would hence still suffer from efficiency problem.

\subsection{Support Neighbor Loss}
We follow the batch sampling strategy of \cite{hermans2017defense} by randomly selecting $P$ person identities, with $Q$ image for each person in a batch. Therefore, in each mini-batch $\mathcal{M}$, we have $M = P*Q$ images.
Taking each $\textbf{x}_i\in\mathcal{M}$ as the anchor, instead of constructing pairs or triplets, we find its $K$-NN neighbors $\mathcal{K}_i=\mathcal{P}_i\cup\mathcal{N}_i$, which consists of positive samples $\mathcal{P}_i$ and negative samples $\mathcal{N}_i$.
It is worth to be noted that samples within $\mathcal{K}_i$ are most similar to $\textit{x}_i$, thus containing the most informative patterns specific to the identity $\textbf{x}_i$ represents.
Samples not included in $\mathcal{K}_i$ are less informative and are simply
discarded for the sake of efficiency and in case that they dilute the contribution of the highly informative ones.
With respect to $\textbf{x}_i$, $\mathcal{P}_i$ and $\mathcal{N}_i$ are the most similar samples with each other, but with different labels, and we need to separate them. In this sense, they are analogous to the ``support vectors'' in SVM. Following this nomenclature, we call them \textit{support neighbors}.

In some sense, we also construct a ``triplet'' $\{\textbf{x}_i, \mathcal{P}_i, \mathcal{N}_i\}$ for $\textbf{x}_i$. 
But unlike the conventional triplet, in our case, $\mathcal{P}_i$ and $\mathcal{N}_i$ are sample groups, rather than individual samples.
These positive and negative neighbors provide more valuable contextual information and neighborhood structure that are absent from the 
conventional triplet. With the expanded hard positive and negative samples, more representative and discriminative information can be 
encoded into the deep model. Besides, for each anchor, we have only one unique ``triplet'' within a batch, so that stable performance is more likely to be guaranteed. 

\begin{figure}
  \begin{center}
    \includegraphics[width=0.75\linewidth]{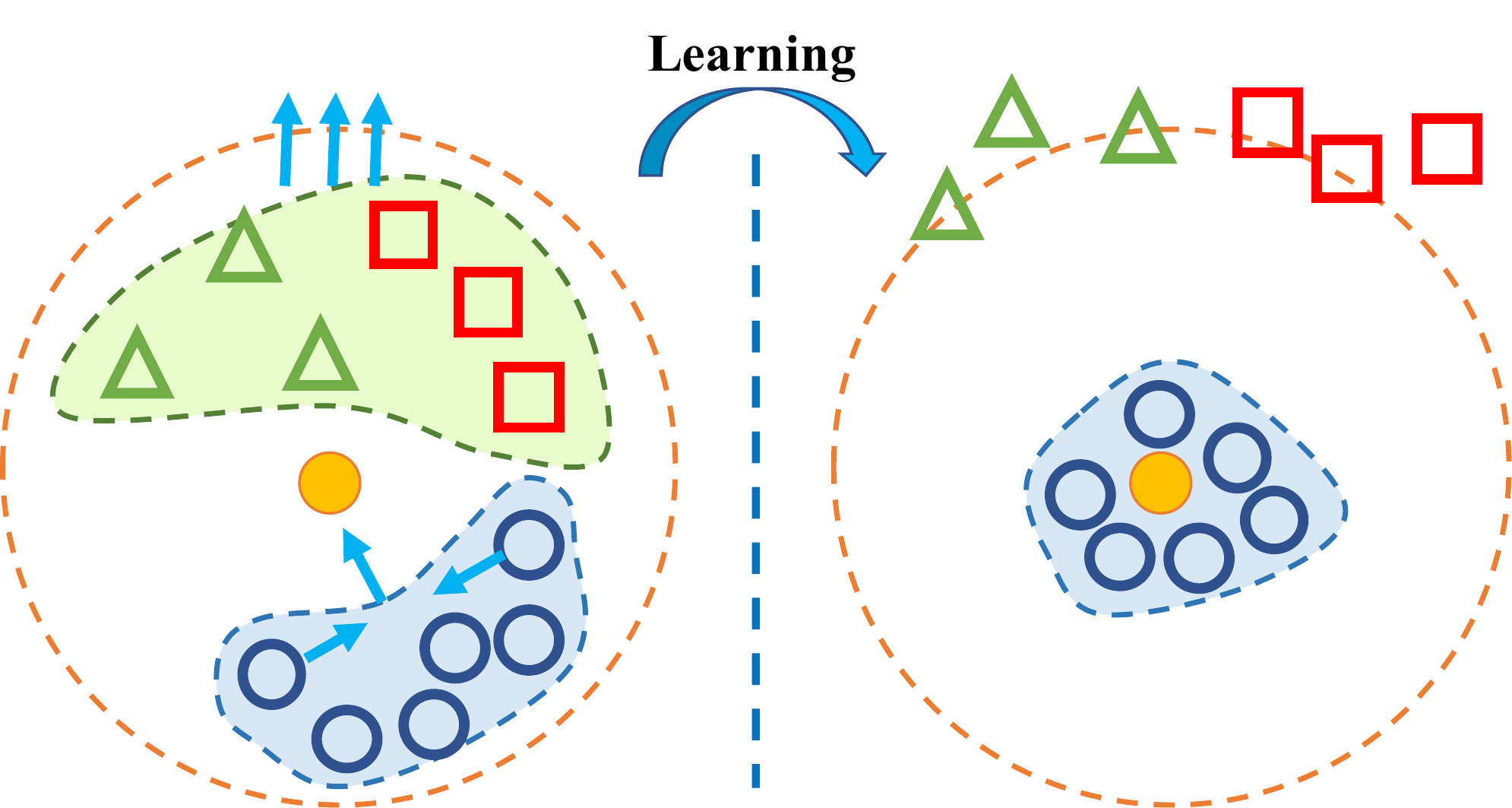}
  \end{center}
  \vspace{-0.10in}
  \caption{Schematic illustration of the proposed support neighbor loss.} 
  \vspace{-0.15in}
  \label{loss_illustration}   
\end{figure}

Based on the support neighbors, we develop the SN loss which aims to separate the positive neighbors from the negative ones, 
and meanwhile penalize the variance among the positive neighbors.
Figure \ref{loss_illustration} shows the schematic illustration of the proposed loss.

\noindent\textbf{Separation loss}.
To separate the positive and negative support neighbors, we seek to maximize the similarity between the anchor $\textbf{x}_i$
and samples from $\mathcal{P}_i$, and meanwhile minimize the similarity between $\textbf{x}_i$ and samples from $\mathcal{N}_i$. 
By this consideration, we define the separation loss as:
\begin{equation}
\begin{array}{cl}
L_{spr}(\theta)= -\sum\limits_{i=1}^N \log\frac{\sum\limits_{\textbf{x}_p \in \mathcal{P}_i} \exp \big(-\sigma D(\textbf{x}_i, \textbf{x}_p)\big)}{
\sum\limits_{\textbf{x}_p \in \mathcal{P}_i} \exp \big(-\sigma D(\textbf{x}_i, \textbf{x}_p)\big) + 
\sum\limits_{\textbf{x}_n \in \mathcal{N}_i} \exp \big(-\sigma D(\textbf{x}_i, \textbf{x}_n)\big)},
\end{array}
\label{separation_loss_long}
\end{equation}
where $\sigma$ is a scaling factor and $D$($\cdot$,$\cdot$) is the euclidean distance between two samples.
We can observe that $S_{ip}=\sum_{\textbf{x}_p \in \mathcal{P}_i} \exp \big(-\sigma D(\textbf{x}_i, \textbf{x}_p)\big)$ 
measures the similarity of $\textbf{x}_i$ to all positive neighbors from $\mathcal{P}_i$, while $S_{in}=\sum_{\textbf{x}_n \in \mathcal{N}_i} \exp \big(-\sigma D(\textbf{x}_i, \textbf{x}_n)\big)$ measures the similarity of $\textbf{x}_i$ to all negative neighbors from $\mathcal{N}_i$.
Since $\mathcal{P}_i\cup\mathcal{N}_i=\mathcal{K}_i$, we can rewrite Eq.\eqref{separation_loss_long} as 
\begin{equation}
\begin{array}{cl}
L_{spr}(\theta)= -\sum\limits_{i=1}^N \log\frac{\sum\limits_{\textbf{x}_p \in \mathcal{P}_i} \exp \big(-\sigma D(\textbf{x}_i, \textbf{x}_p)\big)}
{\sum\limits_{\textbf{x}_s \in \mathcal{K}_i} \exp \big(-\sigma D(\textbf{x}_i, \textbf{x}_s)\big)}
\end{array}
\label{separation_loss_short}
\end{equation}

One may have noticed that our separation loss formulation in Eq. \eqref{separation_loss_short} has a very similar form as the softmax loss formulation in  Eq. \eqref{softmax_loss}. Besides this, they serve the same function to separate different class. 
In fact, if we view $\exp\big(-\sigma D(\textbf{x}_i, \textbf{x}_p)\big)$ as the possibility that $\textbf{x}_i$ and $\textbf{x}_p$ belong to the same class, Eq. \eqref{separation_loss_short} and Eq. \eqref{softmax_loss} are equivalent.

\noindent\textbf{Squeeze loss}.
This loss is to penalize the variance among positive samples, and ``squeeze'' the positive neighbors towards the anchor to form a compact cluster.
It is defined as:
\begin{equation}
\begin{array}{cl}
L_{sqz}(\theta)= \sum\limits_{i=1}^N\Big(\max \big(\underset{\textbf{x}_p\in\mathcal{P}_i}{\bigcup} D(\textbf{x}_i, \textbf{x}_p) \big) - \min \big(\underset{\textbf{x}_p\in\mathcal{P}_i}{\bigcup} D(\textbf{x}_i, \textbf{x}_p) \big)\Big), \\
\end{array}
\label{squeeze_loss}
\end{equation}
where $\underset{\textbf{x}_p\in\mathcal{P}_i}{\bigcup} D(\textbf{x}_i, \textbf{x}_p)$ is the set of distances of every positive neighbor from $\mathcal{P}_i$ to the anchor $\textbf{x}_i$.
Basically, we want to penalize the difference of the distance of an anchor to its furthest positive sample and the distance of the anchor to its nearest positive sample.
In this way, a compact cluster of positive samples can be formed around the anchor point.

During training, the two losses are jointly optimized:
\begin{equation}\label{loss}
L_{sn}(\theta) = L_{spr}(\theta) + \lambda L_{sqz}(\theta).
\end{equation}
where $\lambda$ is the balancing parameter.

To solve the loss function of Eq. \eqref{loss}, we can follow standard deep learning models by adopting Stochastic gradient descent. Specifically, we need to calculate the gradient of $L_{sn}(\theta)$ w.r.t. $\theta$ and we take the $i$-th training sample $\textbf{x}_i$ as an example shown as:
\begin{equation}
\begin{array}{rcl}
\frac{\partial L_{sn}(\theta)}{\partial\theta} & = &  \frac{\partial L_{spr}(\theta)}{\partial\theta} + \lambda \frac{\partial L_{sqz}(\theta)}{\partial\theta} \\
& = & \Big(\frac{\partial L_{spr}(\theta)}{\partial \textbf{x}_i}  + \lambda  \frac{\partial L_{sqz}(\theta)}{\partial \textbf{x}_i}\Big)\frac{\partial \textbf{x}_i}{\partial \theta}
\end{array}
\end{equation}
where the key parts are $\frac{\partial L_{spr}(\theta)}{\partial \textbf{x}_i}$ and $\frac{\partial L_{sqz}(\theta)}{\partial \textbf{x}_i}$, and the detailed derivatives can be found in the the Appendix. $\frac{\partial \textbf{x}_i}{\partial \theta}$ is computed using standard backpropagation.





\subsection{Implementation Details}
Our framework is implemented based on the PyTorch deep learning library. The hardware environment is a PC with Intel Core CPUs (3.6GHz), 48 GB memory, and an NVIDIA GTX TITAN X GPU. Unless specifically noted otherwise, we use the same model structure and training procedure across all experiments and on all datasets.

\noindent\textbf{Training}. We use the ResNet50 architecture as our base network and extract feature of 2048 dimensions for each pedestrian image. The weights of the base network are initialized from the model provided by He \textit{et al.} \cite{he2016deep}. 
For all the datasets, the images are resized to 256$\times$128 and are horizontally flipped to augment training samples. 
We set batch size as 128, with 4 randomly selected images for every 32 identities. 
We adopt the similar exponentially learning rate decaying schedule as \cite{hermans2017defense}. Specifically, we remain the learning rate $r_t$ unchanged as the base learning rate $r_0=2e^{-4}$ until the training epoch reaches $T_s$. After that, we decay the learning rate in the $t$-th epoch as $r_t=r_0*0.001*\frac{t-T_s}{T_f-T_s}$, where $T_f$ is the total training epoch. In our experiments, we set $r_t=75$ and $T_f=800$. We choose the Adam optimizer with the default hyper-parameter values for our experiments.  







\noindent\textbf{Testing}. In the testing stage, we 
feed all the testing images to the CNN model to get their feature embeddings. 
Then we normalize the embeddings to unit vectors. Finally, we compute the Euclidean distances between the embedding vector of each query image and those of every gallery images, and the query images are ranked accordingly.  





\section{Experiments}
\label{experiments}
We validate the effectiveness of the proposed method on three widely used re-ID datasets, namely,
Market-1501 \cite{zheng2015scalable}, CUHK03 \cite{li2014deepreid} and CUHK01 \cite{li2012human}.
We follow the same evaluation protocol as the previous papers \cite{li2014deepreid,ahmed5improved} and 
evaluate the performance by the cumulated matching characteristics (CMC) curve, which is an estimate of the expectation of finding the correct match in the top $n$ matches. 
For Market-1501, the mean average precision (mAP) scores are also reported.

\begin{table}
	\small	
	\renewcommand{\tabcolsep}{4pt}
	\begin{center}
		\caption{Performance comparison on Market-1501. Both single query (SQ) and multiple query (MQ) are evaluated.}				
		\label{market1501}
		\vspace{-10pt}
		\begin{tabular}{|l|c|c|c|c|}\hline			
							  								& \multicolumn{2}{|c|}{SQ}  & \multicolumn{2}{|c|}{MQ} \\\cline{2-5}				
															& mAP & rank-1  & mAP & rank-1  \\\hline			
			LDNS \cite{zhang2016learning} (CVPR16)  & 29.87 & 55.43  & 46.03 & 71.56  \\
			GS-CNN \cite{varior2016gated} (ECCV16)	& 39.55 & 65.88  & 48.45 & 76.04  \\
			CAN  \cite{liu2017end} (TIP17)    & 35.9  & 60.3   & 47.9  &72.1  \\
			JLML \cite{li2017person} (IJCAI17)  & 65.5 & 85.1  & 74.5 & 89.7  \\
			DLCE \cite{zheng2017discriminatively} (TOMM17) & 59.87 & 79.51  & 70.33 &85.84  \\
			LatParts \cite{li2017learning} (CVPR17) & 57.53 & 80.31  & 66.70 & 86.79  \\		
			SSM \cite{bai2017scalable} (CVPR17) & 68.80 & 82.21  & 68.80 & 88.18  \\
			DLPAR \cite{zhao2017deeply} (ICCV17) & 63.4 & 81.0  & - & -  \\
			LSRO \cite{zheng2017unlabeled} (ICCV17) & 56.23 & 78.06 & 56.23 & 85.12  \\
			PDC\cite{su2017pose} (ICCV17) & 63.41 & 84.14  & - & -  \\
			IDE-ML \cite{zhong2017re} (CVPR17) & 49.05 & 73.60 & - & -   \\						
			TriNet \cite{hermans2017defense}  (arxiv17) & 69.14 & 84.92  & 76.42 & 90.53  \\						
			DML \cite{Zhang2018Deep} (CVPR18) & 68.83 & 87.73 & 77.14 & 91.66 \\ 			
			CSA \cite{zhong2018camera}(CVPR18) & 68.72 & 88.12	& - & -	 \\	\hline 
			Ours & \textbf{73.43}  & \textbf{88.27} & \textbf{80.26} & \textbf{92.13} \\  \hline\hline				
			IDE-ML + re-ranking \cite{zhong2017re} (CVPR17) & 63.63 & 77.11 & - & -   \\	
			TriNet + re-ranking \cite{hermans2017defense}  (arxiv17) & 81.07 & 86.67  & 87.18 & 91.75  \\
			CSA + re-ranking \cite{zhong2018camera} (CVPR18) & 71.55 & 89.49	& - & - \\ \hline
			Ours + re-ranking  & \textbf{86.16} & \textbf{89.90} & \textbf{90.27} & \textbf{93.68} \\
			\hline
		\end{tabular}
	\end{center}
			\vspace{-15pt}
\end{table}

\subsection{Comparison with State-of-the-Art}

\noindent\textbf{Market-1501}. 
This dataset includes images of 1,501 persons captured from 6 different cameras. The pedestrians are cropped with bounding-boxes predicted by DPM detector. The whole dataset is divided into training set with 12,936 images of 751 persons and testing set with 3,368 query images and 19,732 gallery images of 750 persons. There are single-query and multiple-query modes in evaluation,
the difference of which is the number of images from the same identity. In multiple-query mode, all features extracted from the images of a person captured by the same camera are merged by average or max pooling, which contains more complete information than single query mode with only 1 query image.

The results for both single-query and multiple-query are shown in Table \ref{market1501}.
We can observe that our method gains the highest mAP and rank-1 accuracy among the methods, for both single-query and multi-query cases. Specially, our method achieves about 5\% performance gains relative to the most recent methods CSA 
\cite{zhong2018camera} and DML \cite{Zhang2018Deep} for the single-query case, while more than 3\% gains relative to DML \cite{Zhang2018Deep} for the multi-query case.
Another remarkable comparison lies between our method and TriNet \cite{hermans2017defense}, 
which also focuses on improving the loss function to achieve discriminative deep feature embeddings from pedestrian images.
Since our method and TriNet use the same base network (Resnet50) to extract features, 
the advantage of our method over TriNet for both the single and multi-query case clearly indicates 
the superiority of our developed loss function. Note that IDE-ML \cite{zhong2017re} is based on the Softmax loss and our SN significantly outperforms it, which shows the advantage of SN loss over traditional identification loss for re-ID.  



We further employ the re-ranking method proposed by Zhong \textit{et al.} \cite{zhong2017re} to improve the obtained results.  The re-ranked results show that our advantages over the most competitive rivalry methods turn even more significant. This further substantiates the superiority of the proposed method.

\begin{table}
	\small
	\renewcommand{\tabcolsep}{4pt}
	\begin{center}
		\caption{Performance comparison on CUHK03. Both labeled and detected pedestrian bounding boxes are experimented.}				
		\label{cuhk03}
			\vspace{-10pt}
		\begin{tabular}{|l|c|c|c|c|}\hline			
			& \multicolumn{2}{|c|}{Labeled}  & \multicolumn{2}{|c|}{Detected} \\\cline{2-5}				
			& rank-1 & rank-5  & rank-1 & rank-5  \\\hline																					
			LDNS \cite{zhang2016learning} (CVPR16)  & 62.55 & 90.05 & 54.70 & 84.75  \\		
			GS-CNN \cite{varior2016gated} (ECCV16)	& - & -         & 61.8 & 86.7  \\
 			CAN  \cite{liu2017end} (TIP17)    & 77.6 & 95.2 & 69.2 & 88.5  \\
 			LatParts \cite{li2017learning} (CVPR17) & 74.21 & 94.33 & 67.99 & 91.04  \\		
			SpindleNet \cite{zhao2017spindle} (CVPR17)  & 88.5 & 97.8 & - & - \\			
			JLML \cite{li2017person} (IJCAI17)  & 83.2 & 98.0 & 80.6 & 96.9  \\
			DLCE \cite{zheng2017discriminatively} (TOMM17) & - & -  & 83.4 & 97.1  \\
			DLPAR \cite{zhao2017deeply} (ICCV17) & 85.4 & 97.6  & 81.6  & 97.3  \\
			SSM \cite{bai2017scalable} (CVPR17) & 76.63 & 94.59 & 72.70 & 92.40  \\
			LSRO \cite{zheng2017unlabeled} (ICCV17) & 84.6 & 97.6 & - & -  \\
			PDC\cite{su2017pose} (ICCV17) & 88.70 & 98.61 & 78.29 & 94.83  \\
			IDE-ML \cite{zhong2017re} (CVPR17) & 61.6 &-  & 58.5 &-   \\
			TriNet \cite{hermans2017defense}  (arxiv17) & 89.6 & \textbf{99.0} & 87.6 & \textbf{98.2}  \\\hline	
			Ours & \textbf{90.2}   & 98.8  & \textbf{88.0}   & 97.7  \\
			\hline
		\end{tabular}
	\end{center}
			\vspace{-15pt}
\end{table}

\noindent\textbf{CUHK03}.
The CUHK03 dataset was collected by two surveillance cameras capturing 14,096 images
of 1,467 persons, with 4.8 images for each person on average. 
The dataset provides both the automatically cropped bounding boxes with a pedestrian detector (Detected) and the manually cropped bounding boxes (Labeled). 
We use both labeled and detected bounding boxes for experiments. Following the testing protocol in \cite{li2014deepreid}, we randomly divide the identities in the dataset into non-overlapping training and test sets, with 1,367 persons for training and 100 persons for testing.
We use the provided 20 partitions for generating training and test sets. 
During the testing stage, for each person, we randomly pick up one image from one camera view as the probe and another image from the other camera view as the gallery. 
The reported cumulative matching characteristic (CMC) are averaged by these 20 groups. Table \ref{cuhk03} shows the proposed method achieves comparable results with TriNet \cite{hermans2017defense}, while surpassing the rest methods with large gaps, especially for the case with detected bounding boxes.



\begin{table}
	\small
	\renewcommand{\tabcolsep}{4pt}
	\begin{center}
		\caption{Performance comparison on CUHK01 with 486 test IDs.}				
		\label{cuhk01_486}
		\vspace{-10pt}
		\begin{tabular}{|l|c|c|c|}\hline			
									    & rank-1 & rank-5  & rank-10 \\\hline																										
			TCP-CNN \cite{cheng2016person} (CVPR16) & 53.7 & 84.3 & 91.0 \\
			LDNS \cite{zhang2016learning} (CVPR16)  & 69.09 & 86.87 & 91.77  \\		
			DCSL \cite{zhang2016semantics} (IJCAI16) & 76.5 & \textbf{94.2} & \textbf{97.5} \\ 
			SSSVM \cite{zhang2016sample} (CVPR16) & 66.0 & 89.1 & 92.8		\\		
			GOG \cite{matsukawa2016hierarchical} (CVPR16) & 57.8 & 79.1 & 86.2 \\ 		
			WARCA \cite{jose2016scalable} (ECCV16) & 65.6 & 85.3 & 90.5 \\ 
 			CAN  \cite{liu2017end} (TIP17)    & 67.2 & 87.3 & 92.5  \\
 			JLML \cite{li2017person} (IJCAI17) & 69.8 & 88.4 & 93.3   \\
 			DLPAR \cite{zhao2017deeply} (ICCV17) & 75.0 & 93.5 & 95.7  \\ \hline
			Ours & \textbf{79.3}  & 94.0 & 97.2   \\\hline
		\end{tabular}
	\end{center}
			\vspace{-15pt}
\end{table}

\noindent\textbf{CUHK01}.
The CUHK01 dataset consists of 3,884 pedestrian images of 971 persons captured by two surveillance cameras. Each person has 4 images, 2 for each camera view. There are two experimental settings, one with 486 identifies for testing and the other with 100 identities for testing. We experiment with both settings.

For the first setting, since there are only a small number of training identities, our network would overfit the training data with high possibility, if directly trained on them. To avoid this, we adopt the strategy of existing methods \cite{ahmed5improved,zhao2017deeply} by firstly training our model with the CUHK03 dataset and then finetuning the model with the 485 training identities. The results in Table \ref{cuhk01_486} show that our method obtains the highest overall matching accuracy, exceeding the 
most competitive algorithm DCSL \cite{zhao2017deeply} by 2.8\% in rank-1 accuracy, while being only slightly inferior for the rank-5 and 10 matching accuracies.

For the second setting, the 851 training identities are sufficient enough for our method to avoid the overfiting problem. We directly train our model on the training data.
We can see from Table~\ref{cuhk01_100} our method achieves very high matching accuracy and is generally better than those of the existing methods. Besides directly training on the 851 training identities, we also train a model finetuned from the pretrained model obtained from the CUHK03 dataset, same as we did for the first setting. The results in Table~\ref{cuhk01_486} show that even high performance is achieved.



\begin{table}
	\small
	\renewcommand{\tabcolsep}{4pt}
	\begin{center}
		\caption{Performance comparison on CUHK01 with 100 test IDs. ``Ours'' refers the results obtained from the model trained from trained data within the dataset. ``Ours (fine-tune)'' denotes that results for the model pretrained from the auxiliary dataset and finetuned with training data within the dataset.}				\vspace{-10pt}
		\label{cuhk01_100}
		\begin{tabular}{|l|c|c|c|}\hline		
									    & rank-1 & rank-5  & rank-10 \\\hline			
			DeepReID \cite{li2014deepreid} (CVPR14)  & 27.9 & 58.2 & 73.5	\\
			IDLA \cite{ahmed5improved} (CVPR15) & 65.0 & 88.7 & 93.1 \\
			Deep Ranking \cite{chen2016deep} (TIP16) & 50.4 & 70.0 & 84.8 \\
			SIR-CIR \cite{wang2016joint} (CVPR16) & 71.8 & 91.6 & 96.0 \\
			PersonNet \cite{wu2016personnet} (ArXiv16) & 71.1 & 90.1 & 95.0 \\
			EDM \cite{shi2016embedding} (ECCV16) & 69.4 & 90.8 & 96.0 \\
			DCSL \cite{zhang2016semantics}  (IJCAI16) & 89.6 & 97.8 & 98.9 \\ 
			DLPAR \cite{zhao2017deeply} (ICCV17) & 88.5 & 98.4 & 99.6  \\ \hline
			Ours & 90.1 & 98.4 & 99.0   \\
			Ours (fine-tune) & \textbf{93.8}  & \textbf{99.0} & \textbf{99.7}   \\\hline
		\end{tabular}
	\end{center}
		\vspace{-15pt}
\end{table}

\subsection{Analytic Study}
\begin{figure}
	\begin{center}
		\includegraphics[width=1.0\linewidth]{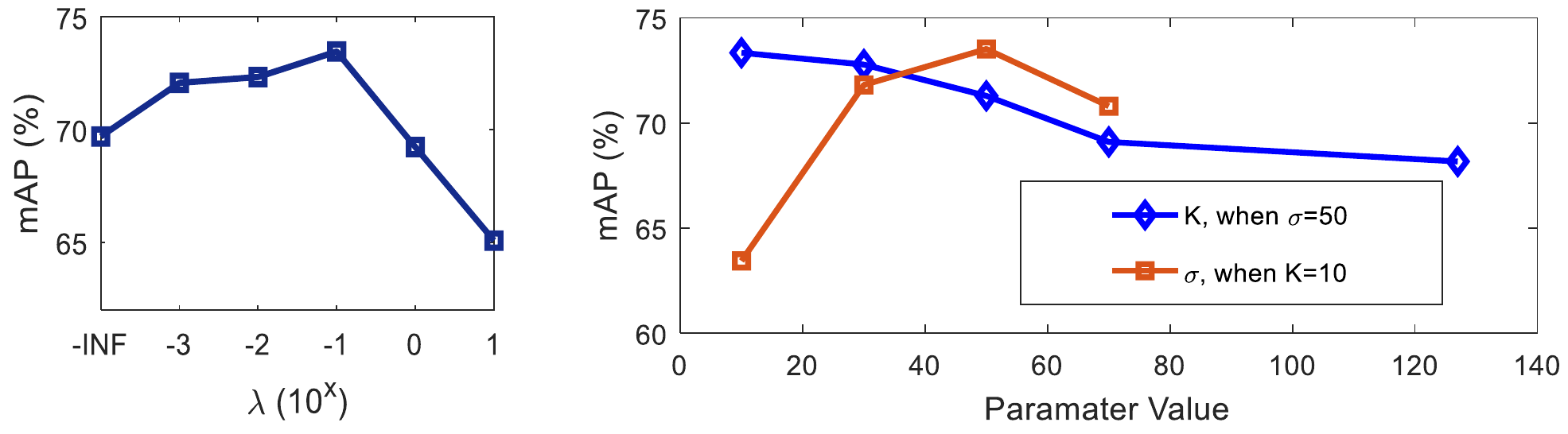}
	\end{center}
	\hspace{-.3in}(a) \hspace{1.5in} (b) 
			\vspace{-10pt}
	\caption{Parameter analysis. (a) the change of mAP value with respect to $\lambda$.
		(b) The change of mAP with respect to parameters $K$ and $\sigma$.} 
	\label{lambda_k_map}	
	 \vspace{-15pt}	
\end{figure}

\noindent\textbf{Parameter Analysis}. The proposed Support Neighbor (SN) loss in Eq. (\ref{loss}) consists of two the components, the separation loss and the squeeze loss, which are balanced by the hyper-parameter $\lambda$. The separation loss serves to enlarge the margins between the positive neighbors and negative neighbors for an anchor, while the squeeze loss aims to penalize the variance among the positive neighbors. To evaluate the impact of the two losses to the performance, we vary $\lambda$ with the values in $\{0, 10^{-3}, 10^{-2}, 10^{-1}, 1, 10\}$ and calculate the mAP value on the Market1501 dataset. Note $\lambda=0$ means the squeeze loss is not considered, and only the separation loss component is used to train the network. The results in the Figure \ref{lambda_k_map}(a) show that when simultaneously considering both the two types of losses, higher mAP can be obtained. This substantiates the squeeze loss indeed contributes to more discriminative embeddings than that using the separation loss alone. We also observe the mAP reaches the peak when $\lambda=0.1$, and thus we use this setting for all the experiments. 

Besides $\lambda$, there are two other hyper-parameters $\sigma$ and $K$ in the proposed method. $\sigma$ is the scaling factor for the separation loss, while $K$ is the number of nearest neighbors selected for calculating the loss. We analyze the impact of the two parameters separately by fixing one while evaluating the other. Figure 
\ref{lambda_k_map}(b) shows the result. We can observe that while fixing $\sigma$, the mAP value drops consistently as we enlarge the value for $K$. For $K=127$, i.e., all the other  samples in a batch are regarded as the neighbors for any given sample, the lowest mAP is obtained. This shows that the incorporation of the less informative samples for calculating the loss may ``washing out'' the contribution of the high informative ones. A similar observation can be found in \cite{hermans2017defense}, where the $\textit{Batch Hard}$ loss (only the hardest triplet is selected for calculating the loss) leads to better performance than the $\textit{Batch All}$ loss (all samples in a batch are used to construct triplets and contribute to the loss). 

As to the scaling factor $\sigma$, a smaller value for it will leads to a big separation loss. The experimental results show that it is favored to set its value bigger than 30.

\begin{figure}
	\begin{center}
		\includegraphics[width=1.0\linewidth]{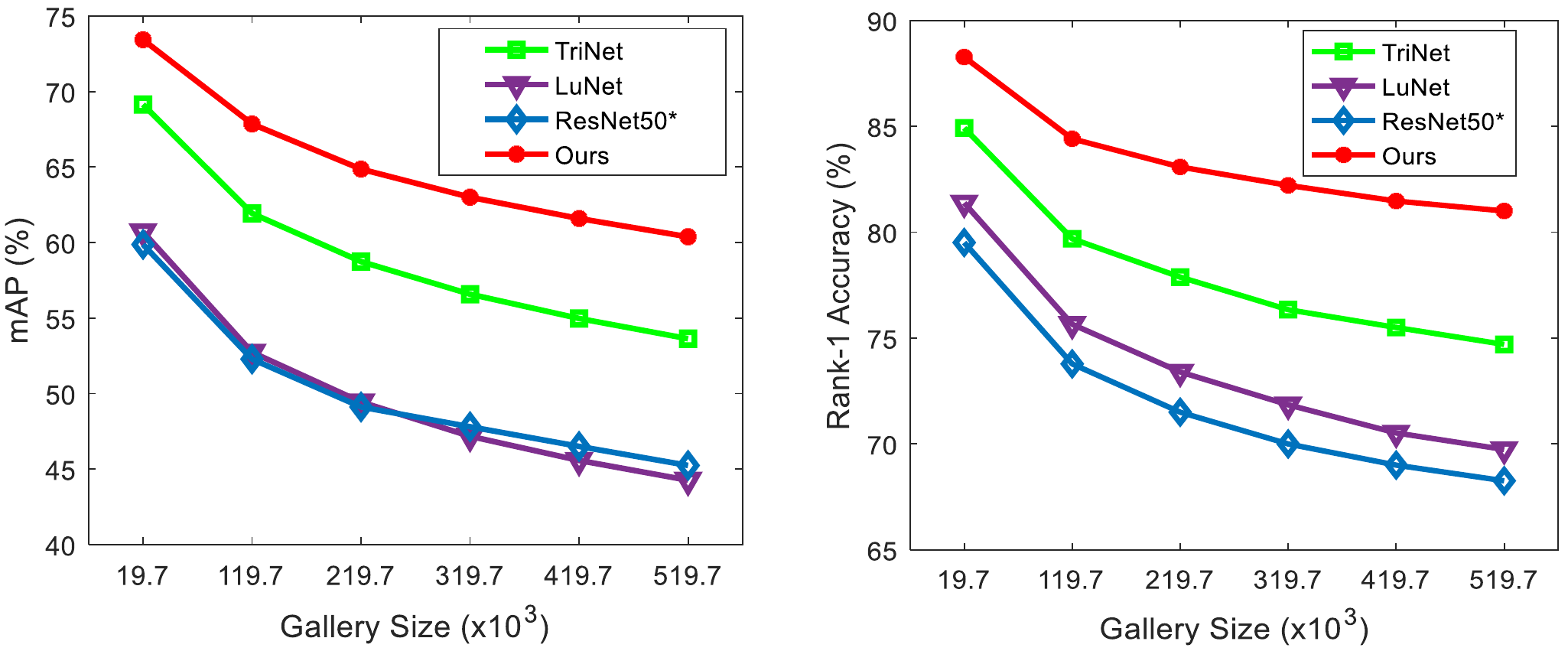}
	\end{center}
	\vspace{-0.15in}
	\caption{Impact of gallery size. Data of the compared methods (namely, TriNet, LuNet and $\textrm{Resnet50}^{*}$) are from \cite{hermans2017defense}. } 
	\vspace{-0.25in}
	\label{gallery_size}		
\end{figure}

\noindent\textbf{Impact of Gallery Size}.
One of the most challenging problems of applying person re-ID in practice is how to secure satisfactory performance when the gallery set is significantly enlarged. To study the performance of the proposed method for addressing this problem, we employ the distractor set provided along with the Market1501 dataset, use it to supply the gallery set and evaluate the re-ID performance with the enlarged gallery set. 
The distractor set includes 500 thousand image bounding boxes, consisting of the persons not belonging to any of the original 1,501 identities and false alarms on the background.
The distractor set is more than 25 times larger than the original gallery set (about 19 thousands), which 
makes the retrieval more difficult and aligns better with realistic setting. 
We progressively add the distractors into the gallery set by random selection. The rank-1 accuracy and mAP value of our method and several existing ones are shown in Figure \ref{gallery_size}. We can observe that the performances of all the methods degenerate as the gallery set is enlarged. Comparatively speaking, our method consistently maintains the best performance with all gallery sizes, and there is a clear trend that our advantage becomes even more significant as the gallery set turns larger.

\begin{figure*}
	\begin{center}
		\includegraphics[width=1.0\linewidth]{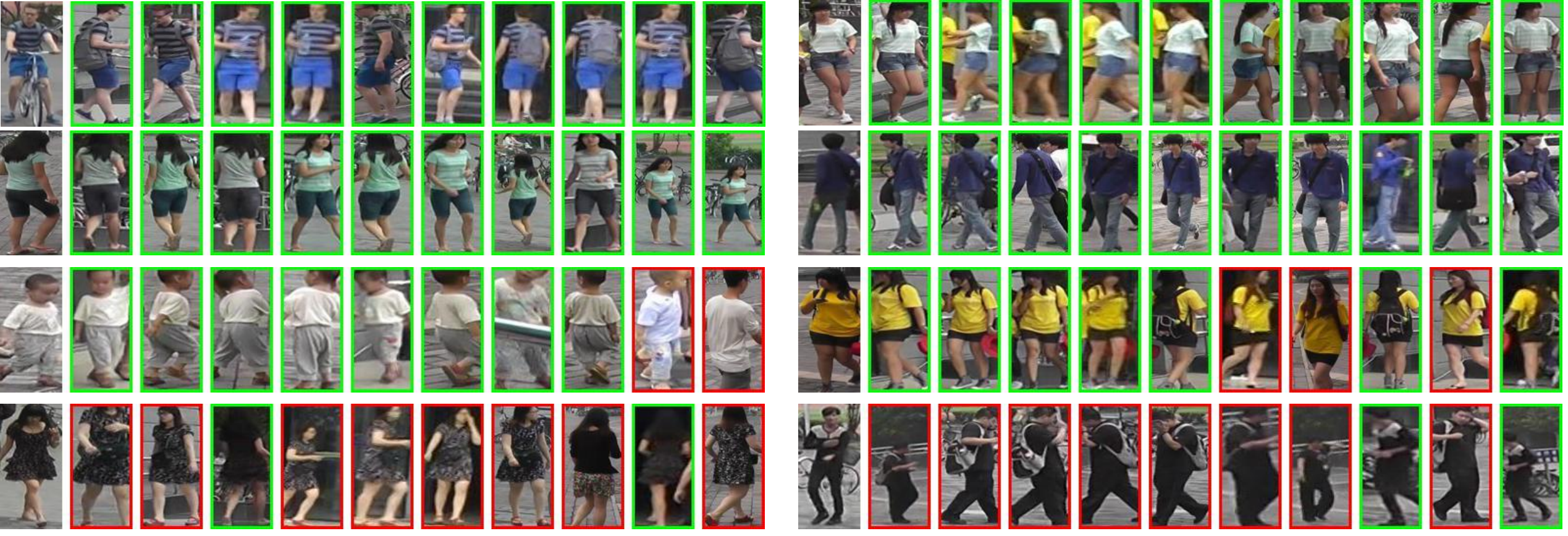}
	\end{center}
	\caption{Sample person retrieval results on Market1501. 
	The most left (without border) in each of the six groups is the query and the rest are the top-10 ranked returns from the gallery. Correct returns are in green borders, while incorrect ones are in red borders.} 
	\label{visualization}		
\end{figure*}

\begin{figure*}
	\begin{center}
		\includegraphics[width=1.0\linewidth]{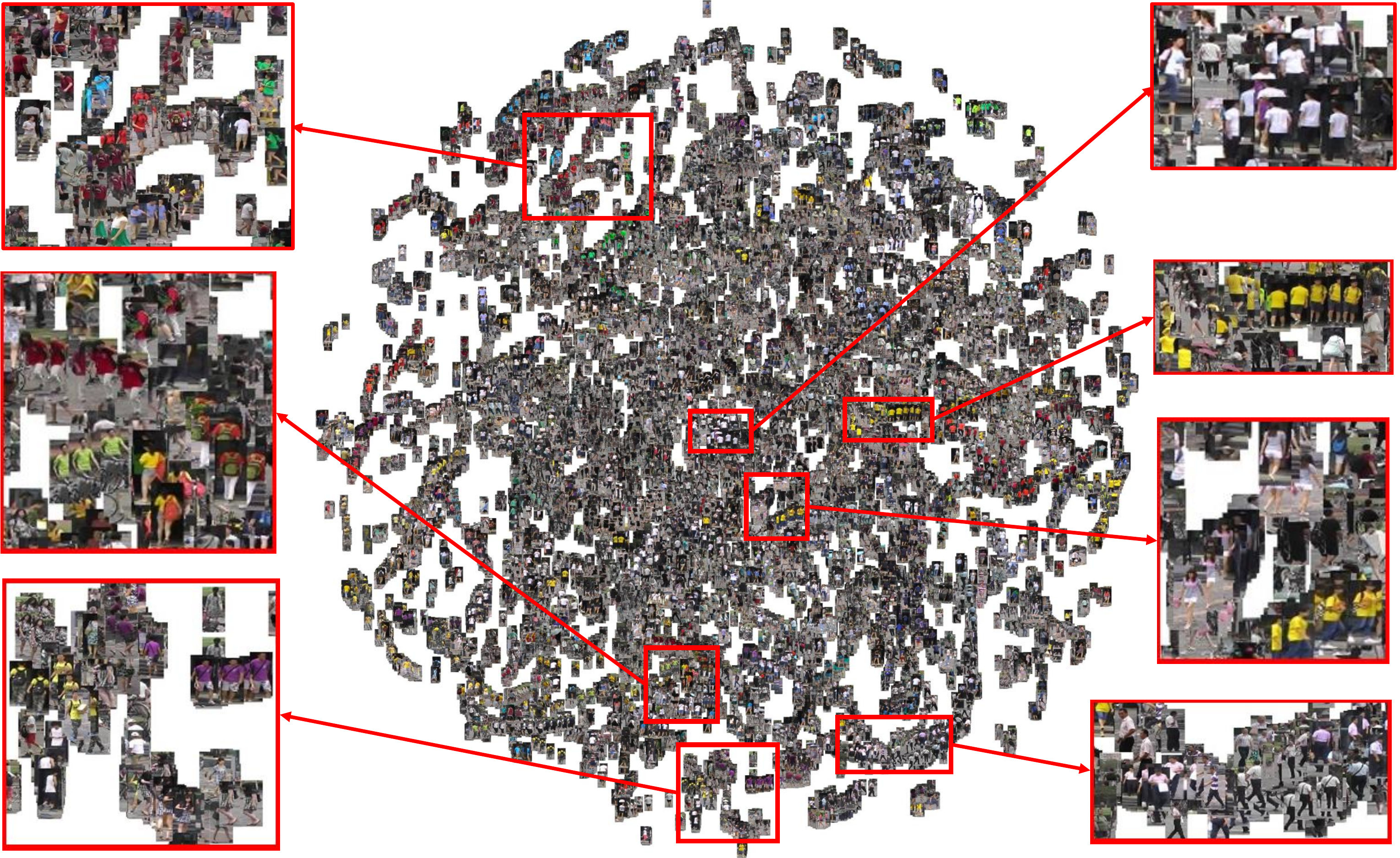}
	\end{center}
	\caption{The Barnes-Hut t-SNE \cite{van2014accelerating} of the learned embeddings for the test set of the Market-1501 dataset. We can observe that satisfactory embeddings have been achieved and images of the same identities are closely distributed.} 
	\label{embedding}		
\end{figure*}

\noindent\textbf{Result Visualization}.
Figure \ref{visualization} shows some retrieval results on the Market1501 dataset.
We can observe that the proposed method is fairly robust to get the correct retrieval results
and fails only in some very hard cases that are challenging even for human.
Figure \ref{embedding} shows the learned embedding of the test set of the Market1501 dataset. We can
see that images of the same person are closely distributed for the majority of the case. This substantiates
the effectiveness of the proposed loss on guiding to learn effective deep embeddings.

\section{Conclusion}
\label{conclusions}
This paper proposed a new loss function for person re-identification, called Support Neighbor (SN) loss, which comprises of the separation component and squeeze component.
The separation component serves to push part the positive neighbors from the negative ones of every anchor point within a mini-batch,
while the squeeze component pushes together positive neighbors of the anchor to form a compact cluster.
We verify the effective of SN loss by integrating it on top of Resnet50, and the experiments show that we achieve the state-of-the-art on several benchmarking datasets.
Experiments also show that the two components of SN loss complement with each other, and the advantage of SN loss becomes more significant compared with existing losses when the gallery dataset is tremendously enlarged.


\begin{acks}
This research is supported in part by the NSF IIS Award 1651902, ONR Young Investigator Award N00014-14-1-0484, and U.S. Army Research Office Award W911NF-17-1-0367.
\end{acks}











\section*{Appendix}

In this appendix, we show the key steps of $\frac{\partial L_{spr}(\theta)}{\partial \textbf{x}_i}$ and $\frac{\partial L_{sqz}(\theta)}{\partial \textbf{x}_i}$.

First of all, we define $d_{ip} = \sum\limits_{\textbf{x}_p \in \mathcal{P}_i} \exp \big(-\sigma D(\textbf{x}_i, \textbf{x}_p)\big)$ and $d_{is} = 
\sum\limits_{\textbf{x}_s \in \mathcal{K}_i} \exp \big(-\sigma D(\textbf{x}_i, \textbf{x}_s)\big)$. $D(\textbf{x}_i, \textbf{x}_p) = \|f(\textbf{x}_i|\theta)-f(\textbf{x}_p|\theta)\|_2^2$ is the Euclidean distance for outputs of sample $\textbf{x}_i$ and $\textbf{x}_p$. 
For simplicity, we further define $\kappa = \frac{d_{ip}}{d_{is}}$, thus we have $\frac{\partial L_{spr}(\theta)}{\partial \textbf{x}_i} =-\frac{1}{\kappa}\frac{\partial \kappa}{\partial \textbf{x}_i}$ and for $\frac{\partial \kappa}{\partial \textbf{x}_i}$, we need to consider three different cases for $\textbf{x}_i$. 

For the first case:
\begin{equation}
\vspace{-5pt}
\begin{array}{rcl}
\frac{\partial \kappa}{\partial \textbf{x}_i} &  = & \frac{\frac{\partial d_{ip}}{\partial \textbf{x}_i}d_{is}- \frac{\partial d_{is}}{\partial \textbf{x}_i}d_{ip}}{d_{is}^2} \\
&  = & \kappa\delta\Big(-\sum\limits_{\textbf{x}_s \in \mathcal{K}_i}\frac{\partial D(\textbf{x}_i,\textbf{x}_s)}{\partial \textbf{x}_i}+\sum\limits_{x_p \in \mathcal{P}_i}\frac{\partial D(\textbf{x}_i,\textbf{x}_p)}{\partial x_i}\Big) \\
&  = & 2\kappa\delta\Big(-\sum\limits_{\textbf{x}_s \in \mathcal{K}_i}(\textbf{x}_i-\textbf{x}_s)+\sum\limits_{\textbf{x}_p \in \mathcal{P}_i}(\textbf{x}_i-\textbf{x}_p)\Big) \\
\end{array}
\end{equation}

For the second case, we consider other sample $\textbf{x}_q$ find $\textbf{x}_i$ as its positive pair. Then we $d_{pi} = \sum\limits_{\textbf{x}_i \in \mathcal{P}(\textbf{x}_q)} \exp \big(-\sigma D(\textbf{x}_q, \textbf{x}_i)\big)$ and $d_{si} = \sum\limits_{\textbf{x}_i \in \mathcal{K}(\textbf{x}_q)} \exp \big(-\sigma D(\textbf{x}_q, \textbf{x}_i)\big)$:
\begin{equation}
\vspace{-5pt}
\begin{array}{rcl}
\frac{\partial \kappa}{\partial \textbf{x}_i} &  = & \frac{\frac{\partial d_{pi}}{\partial \textbf{x}_i}d_{si}- \frac{\partial d_{si}}{\partial \textbf{x}_i}d_{pi}}{d_{si}^2} \\
&  = & \kappa\delta\Big(-\sum\limits_{\textbf{x}_i \in \mathcal{K}(\textbf{x}_q)}\frac{\partial D(\textbf{x}_q,\textbf{x}_i)}{\partial \textbf{x}_i}+\sum\limits_{\textbf{x}_i \in \mathcal{P}(\textbf{x}_q)}\frac{\partial D(\textbf{x}_q,\textbf{x}_i)}{\partial \textbf{x}_i}\Big) \\
  &  = & 2\kappa\delta\Big(-\sum\limits_{\textbf{x}_i \in \mathcal{K}(\textbf{x}_q)}(\textbf{x}_i-\textbf{x}_q)+\sum\limits_{\textbf{x}_i \in \mathcal{P}(\textbf{x}_q)}(\textbf{x}_i-\textbf{x}_q)\Big) \\
\end{array}
\end{equation}
where $\mathcal{K}(\textbf{x}_q)$ is the set of $K$-nearest neighbors to $\textbf{x}_q$ and $\mathcal{P}(\textbf{x}_q)$ is the set of positive neighbors to $\textbf{x}_q$.

For the third case, we consider other sample $\textbf{x}_q$ find $\textbf{x}_i$ as its neighbors but not positive pair.Then we $d_{pi} = \sum\limits_{\textbf{x}_i \notin \mathcal{P}(\textbf{x}_q)} \exp \big(-\sigma D(\textbf{x}_q, \textbf{x}_i)\big)$, which is a constant and $d_{si} = \sum\limits_{\textbf{x}_i \in \mathcal{K}(\textbf{x}_q) ~\&~ \textbf{x}_i \notin \mathcal{P}(\textbf{x}_q)} \exp \big(-\sigma D(\textbf{x}_q, \textbf{x}_i)\big)$:
\begin{equation}
\begin{array}{rcl}
\frac{\partial \kappa}{\partial \textbf{x}_i} &  = & \frac{- \frac{\partial d_{si}}{\partial \textbf{x}_i}d_{pi}}{d_{si}^2} = \kappa\delta\Big(-\sum\limits_{\textbf{x}_i \in \mathcal{K}(x_q) ~\&~ \textbf{x}_i \notin \mathcal{P}(\textbf{x}_q)}\frac{\partial D(\textbf{x}_q,\textbf{x}_i)}{\partial \textbf{x}_i}\Big) \\
&  = &  2\kappa\delta\Big(-\sum\limits_{\textbf{x}_i \in \mathcal{K}(\textbf{x}_q) ~\&~ \textbf{x}_i \notin \mathcal{P}(\textbf{x}_q)}(\textbf{x}_i-\textbf{x}_q)\Big).
\end{array}
\end{equation}
Thus, we have $\frac{\partial L_{spr}(\theta)}{\partial \textbf{x}_i} = 2\delta\Big(\sum\limits_{\textbf{x}_s \in \mathcal{K}_i}(\textbf{x}_i-\textbf{x}_s)-\sum\limits_{\textbf{x}_p \in \mathcal{P}_i}(\textbf{x}_i-\textbf{x}_p)\Big) + 2\delta\Big(\sum\limits_{\textbf{x}_i \in \mathcal{K}(\textbf{x}_q)}(\textbf{x}_i-\textbf{x}_q)-\sum\limits_{\textbf{x}_i \in \mathcal{P}(\textbf{x}_q)}(\textbf{x}_i-\textbf{x}_q)\Big) + 2\delta\Big(\sum\limits_{\textbf{x}_i \in \mathcal{K}(\textbf{x}_q) ~\&~ \textbf{x}_i \notin \mathcal{P}(\textbf{x}_q)}(\textbf{x}_i-\textbf{x}_q)\Big)$.

For $\frac{\partial L_{sqz}(\theta)}{\partial\theta}$, we need also consider three cases, i.e., $\textbf{x}_i$ is the anchor, $\textbf{x}_j$ is the anchor and $\textbf{x}_i$ becomes the closest or farthest sample: 
\begin{equation}
\begin{array}{rcl}
\frac{\partial L_{sqz}(\theta)}{\partial\theta} &  = & 2(\textbf{x}_i-x_f)-2(\textbf{x}_i-\textbf{x}_c) + \sum\limits_{\textbf{x}_i = \mathcal{P}(\textbf{x}_j,f) }2(\textbf{x}_i-x_j)  \\
												&    & +\sum\limits_{\textbf{x}_i = \mathcal{P}(\textbf{x}_j,c) }2(\textbf{x}_i-\textbf{x}_j) \\												
												&  = & 2\textbf{x}_c-\textbf{x}_f + \sum\limits_{\textbf{x}_i = \mathcal{P}(\textbf{x}_j,f) }2(\textbf{x}_i-\textbf{x}_j) \\
												&    & +\sum\limits_{\textbf{x}_i = \mathcal{P}(\textbf{x}_j,c) }2(\textbf{x}_i-\textbf{x}_j) \\
\end{array}
\end{equation}
where $\textbf{x}_f$ is the farthest sample while $\textbf{x}_c$ is the closet sample when $\textbf{x}_i$ works as the anchor. $\mathcal{P}(\textbf{x}_j,f)$ is the farthest positive sample of $\textbf{x}_j$ and $\mathcal{P}(\textbf{x}_j,c)$ is the closest positive sample of $\textbf{x}_j$.

So far, we have obtained the derivatives of $\frac{\partial L_{/}(\theta)}{\partial \textbf{x}_i}$ and $\frac{\partial L_{sqz}(\theta)}{\partial \textbf{x}_i}$.

\bibliographystyle{ACM-Reference-Format}

\end{document}